\documentclass[A4paper, 11 pt, conference]{ieeeconf}  
\IEEEoverridecommandlockouts       
\usepackage{geometry}
\usepackage{xcolor}
\usepackage{graphics} 
\usepackage{epsfig} 
\usepackage{mathptmx} 
\usepackage{times} 
\usepackage{amsmath} 
\usepackage{amssymb}  
\usepackage{subfigure}
\usepackage{algorithm} 
\usepackage{algorithmic} 
\usepackage{multirow} 
\begin{document}

\title{\LARGE \bf
Multi-robot Path Planning with Rapidly-exploring Random Disjointed-Trees}

\author{Biru Zhang$^{1,2}$, Jiankun Wang$^{1,2,3}$, \textit{IEEE Senior Member}, Max Q.-H. Meng$^{1,2}$, \textit{IEEE Fellow} 
\thanks{This work is supported by the Shenzhen Key Laboratory of Robotics Perception and Intelligence under Grant ZDSYS20200810171800001, Shenzhen Outstanding Scientific and Technological Innovation Talents Training Project under Grant RCBS20221008093305007, and National Natural Science Foundation of China grant 62103181. \textit{Correpsonding authors: Jiankun Wang and Max Q.-H. Meng.} E-mail: wangjk@sustech.edu.cn, max.meng@ieee.org}
\thanks{$^1$Shenzhen Key Laboratory of Robotics Perception and Intelligence, Shenzhen, China. $^2$Department of Electronic and Electrical Engineering, Southern University of Science and Technology, Shenzhen, China. $^3$Jiaxing Research Institute, Southern University of Science and Technology, Jiaxing, China.}
}

\maketitle 
\thispagestyle{empty}

\begin{abstract}
Multi-robot path planning is a computational process involving finding paths for each robot from its start to the goal while ensuring collision-free operation. 
It is widely used in robots and autonomous driving. 
However, the computational time of multi-robot path planning algorithms is enormous, resulting in low efficiency in practical applications.
To address this problem, this article proposes a novel multi-robot path planning algorithm (Multi-Agent Rapidly-exploring Random Disjointed-Trees*, MA-RRdT*) based on multi-tree random sampling. 
The proposed algorithm is based on a single-robot path planning algorithm (Rapidly-exploring Random disjointed-Trees*, RRdT*). 
The novel MA-RRdT* algorithm has the advantages of fast speed, high space exploration efficiency, and suitability for complex maps. 
Comparative experiments are completed to evaluate the effectiveness of MA-RRdT*. 
The final experimental results validate the superior performance of the MA-RRdT* algorithm in terms of time cost and space exploration efficiency.
\end{abstract}

\section{INTRODUCTION}
Path planning is a vibrant research area in robotics that aims at finding a safe and feasible path in a given environment for a single robot to travel from a start point to a goal point \cite{wang2019finding}. 
The multi-robot path planning focuses on providing collision-free paths for multiple robots to complete tasks in the same environment cooperatively. 
Compared with single-robot path planning, multi-robot path planning is more challenging due to its increased complexity. 
But it also has broader applications. 
As a result, multi-robot path planning is an active and essential research topic in fields such as unmanned aerial vehicles, mobile robots, and cooperative robots \cite{madridano2021trajectory}. 
Consequently, the study of multi-robot path planning holds significant importance in promoting the development of robotics technology.

One of the critical challenges in multi-robot path planning is to obtain an optimal solution efficiently. 
As the number of agents increases, the problem's complexity increases, and the search space grows exponentially. 
This makes finding a feasible solution within reasonable time a challenging task. Therefore, developing novel and efficient algorithms to address this challenge is significant for advancing multi-robot path planning research.

\subsection{Related work}

Traditional path planning algorithms can be mainly divided into two categories, namely graph-based algorithms and sampling-based algorithms. 
Graph search algorithms search the space to find the shortest path. 
These algorithms perform well in cases with small search spaces. 
However, their computational complexity increases significantly when the search space gets larger \cite{yamamoto1998formalization}. 
On the other hand, sampling-based algorithms generate paths via random sampling in the state space. 
The representative algorithm is Rapidly-exploring Random Tree* (RRT*) \cite{karaman2011sampling}. 
These algorithms can utilize randomness, require less computation, and quickly generate feasible paths.

For multi-robot path planning, there are two main methods, the centralized method and the decentralized method. 
A central planner acquires information about the environment and all robots. 
It uses this information to plan paths for each robot \cite{luna2011efficient}. 
These methods typically use graph search algorithms, such as A* \cite{hart1968formal}, and combine them with conflict-based search methods to find collision-free paths for robots, such as the Conflict-Based Search (CBS) algorithm and its variants \cite{sharon2015conflict}, \cite{felner2018adding}, \cite{boyarski2015icbs}. 
However, these algorithms are computationally complex. 
And they can be seriously limited by the space. 
In large maps, these methods can experience significant increases in computation time. On the other hand, the decentralized approach decomposes the problem into multiple sub-problems. 
It allows each robot to independently solve its sub-problem and collaborate to complete the task. 
Each robot needs to find a feasible path for itself. 
Meanwhile, each robot regards other robots as dynamic obstacles and makes decisions based on its own local information \cite{chen2021decentralized}. 
Robots can perceive each other's position information to coordinate their movements. This algorithm mainly consists of global path planning and local collision avoidance. In global path planning, sampling-based algorithms are mainly used to enable each robot to find its feasible path. 
For example, the Multi-agent RRT* (MA-RRT*) algorithm is based on RRT*, an asymptotic optimal sampling algorithm \cite{vcap2013multi}. 
In MA-RRT* algorithm, each robot uses the RRT* algorithm to generate its own path. 
Some decentralized methods utilize the improved version of RRT* as the global path planning module to increase speed or reduce memory \cite{9194248}, \cite{8814874}, \cite{ragaglia2016multi}.

Considering the advantages of the decentralized algorithm's fast speed and low computational complexity, this article adopts the sampling-based path planning algorithm to solve the multi-robot path planning problem. 
This article aims to find the global path in the shortest possible time. 
With the RRdT* algorithm as the basis \cite{8793618}, we call the proposed algorithm MA-RRdT* algorithm. 
Experiment results demonstrate that the proposed algorithm improves the performance of multi-robot path planning significantly.

\section{PRELIMINARIES}

\subsection{RRdT*}

The RRdT* algorithm is a sampling-based path planning algorithm based on RRT*. 
It has the advantages of fast speed and is suitable for complex maps.
RRdT* employs multiple trees to search the space effectively. 
These trees originate from randomly distributed root nodes within the map and are iteratively expanded to explore local connectivity. 
Within the RRdT* algorithm, two primary types of exploration trees exist: the root tree, which originates from the start point, and the disjointed tree, which is randomly generated within the space. 
The algorithm generates a root tree at the start and multiple disjointed trees throughout the map. 
Each tree maintains a probability value that is modified based on the success or failure of samplings. 
Continuously failed samplings result in a decreased probability value, indicating reduced significance within the path planning procedure. 
The algorithm utilizes the restart mechanism of disjointed trees and the chained directed sampling during the sampling process. 
The detail of these two methods is described as follows:

\paragraph{Restart Mechanism of Disjointed Trees} In each sampling iteration, the probabilities of all disjointed trees are evaluated. 
If the probability of a disjointed tree falls below a predefined threshold, the tree is restarted at a randomly selected new location, denoted as $q_{rand}$. 
By allowing the creation of a new tree when there are multiple failed exploration attempts for an existing tree, the RRdT* algorithm effectively addresses the challenge of connection failures between nodes in highly constrained spaces. 
Importantly, if $q_{rand}$ is located near any specific exploration tree, it will become a node of that tree.

\begin{figure}[htbp]
\centerline{\includegraphics{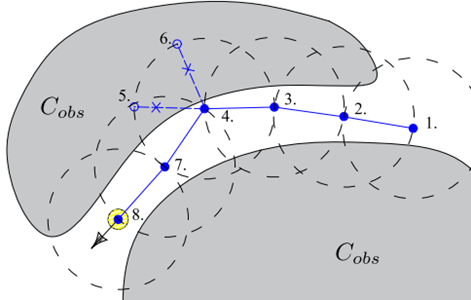}}
\caption{Example of chained directed samples \cite{8793618}. The disjointed tree initializes at location 1 and exhibits a consecutive sequence of successful samples from 2 to 4. The fifth sample fails due to a collision. Using the sampler, the disjointed tree retries the sampling process and obtains sample 6, which also fails. However, the seventh sample successfully identifies a free path for continued chained directed sampling.}
\label{channed}
\end{figure}

\paragraph{Chained Directed Sampling} In the sampling process of the disjointed trees, the algorithm utilizes the Markov Chain Monte Carlo (MCMC) method with the von Mises-Fisher distribution, a type of Gaussian distribution, to obtain information about previously successful samples \cite{wood1994simulation}. 
This mechanism enables the algorithm to generate sequential directed samples through the sampler. 
Fig.~\ref{channed} shows the example of chained directed sampling.

\begin{figure}[htbp]
\centerline{\includegraphics{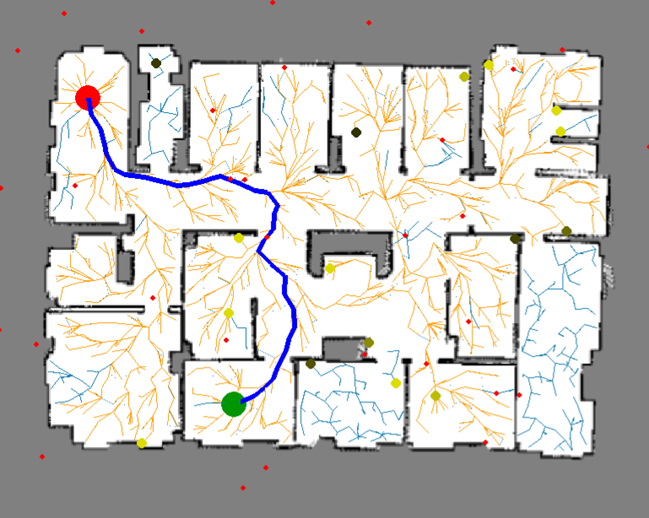}}
\caption{Example of RRdT* algorithm. The red dot corresponds to the start point, while the green dot represents the goal point. The thick dark blue line depicts the planned path, with the narrow orange line portion indicating the root tree and the narrow light blue line portion representing the disjointed tree.}
\label{RRdT}
\end{figure}

Algorithm~\ref{RRdT_a} explains the RRdT* algorithm. 
Initially, $k$ points are randomly picked within the obstacle-free space $C_{free}$ to serve as the root nodes of the disjointed trees. 
Subsequently, a tree (referred to as $t$) is selected according to its probability values. 
By employing the MCMC method, the chosen tree expands its local connectivity. 
Let $q_{new}$ represent a newly obtained sample through the chained directed sampling procedure. 
If $q_{new}$ does not collide with any obstacles, it is added to all trees within a predefined distance. 
This process involves examining all nodes on other trees that are close to $q_{new}$ (i.e., identifying trees near $t$) and merging these trees with $t$. 
If the root tree is sufficiently close to $t$, all nodes of $t$ are integrated into the root tree. 
Additionally, when a new node joins the root tree, it needs to be connected using the edge rewiring process, similar to RRT*. 
This process enables a certain level of path optimization. 
Subsequently, the position of tree $t$ is updated to $q_{new}$, and its probability value is adjusted accordingly. 
This iterative process continues until the desired number of sampled nodes is achieved. Fig.~\ref{RRdT} illustrates the operation process of the RRdT* algorithm.

\begin{algorithm}
    \caption{RRdT*}
    \label{RRdT_a}
    \begin{algorithmic}[1]
        \REQUIRE 1 start points, 1 goal points
        \STATE Initialize one root tree and $k$ disjointed trees
        \WHILE{$n < N$}
            \STATE Restart the disjointed trees with low probability values
            \STATE $t \gets$ Randomly pick a tree
            \STATE $q_{new} \gets t$ samples
            \STATE Merge $t$ with other trees within $\epsilon$-distance of $q_{new}$
            \STATE Update $t$ probability
        \ENDWHILE
    \end{algorithmic}
\end{algorithm}

\section{MA-RRdT*}

This work focuses on improving the RRdT* algorithm to address the multi-robot path planning problem, leading to an enhanced algorithm called MA-RRdT*. 
Each robot has its distinct start and goal, so the corresponding start and goal points are assigned to form a group. 
The algorithm aims to find the paths for each group. 
Four key improvements have been made to the RRdT* algorithm to achieve this goal, resulting in the development of MA-RRdT*.

\subsection{Bidirectional root trees and shared disjointed trees}\label{AA}

The MA-RRdT* algorithm implements bidirectional root trees and shared disjointed trees to ensure efficient path planning for each robot. 
Bidirectional root trees mean that a dedicated root tree is generated for each start and goal. 
Therefore, in scenarios involving $n$ robots, the map contains a total of $2n$ root trees along with several disjointed trees. 
Once a node from a root tree reaches sufficient proximity to a node from the root tree of the same group (and no obstacles obstruct their path), it indicates that a feasible path for that specific group is found. 
Once a path is generated, the corresponding robot starts moving along the path.

In addition, the algorithm shares the location information obtained from exploring the disjointed trees on the map. 
The MA-RRdT* algorithm uses multiple disjointed trees to explore the space. 
And the spatial information obtained by these trees can be accessed by each group of start and goal. 
In the RRdT* algorithm, if a disjointed tree is sufficiently close to a root tree, it will be merged with the root tree (all nodes in the disjointed tree are added to the root tree) and restarted. 
However, in the MA-RRdT* algorithm, when a disjointed tree is merged with the root tree, a new node is generated at each node's position, and these new nodes are joined to the root tree. 
This aims to ensure that a disjointed tree's information can be shared. 
These new nodes carry the same location information as the original nodes. 
After the first merging occurs, the disjointed tree is marked as inactive and stops growing (simultaneously, new disjointed trees are generated at random locations on the map). 
Once another root tree is sufficiently close to the inactive disjointed tree, the location information of all nodes in the tree is shared with the root tree (i.e., new nodes that carry the location information about the disjointed tree are added to the root tree). 
This approach allows the spatial information recorded by the disjointed tree to be shared among all root trees. 
It is important to note that the spatial information of the same disjointed tree can only be added to one of the root trees in each group. 
This ensures that redundant spatial information is not added within the same group of root trees.

\subsection{Heuristic method}

Due to variations in the positions of start and goal points, some groups find paths slower than others. 
To expedite the path planning process for all groups and ensure efficient path planning for each robot, the algorithm introduces a heuristic method to minimize the algorithmic run time. 
Once a group of root trees successfully finds a path, the probability values of the trees in that group are set to 0. 
Moreover, when identifying trees close to a newly sampled node $q_{new}$, the root trees that have found their path are excluded. 
This prevents the root trees in this group from sampling and merging with disjointed trees within the space. 
Consequently, growth is stopped for the root trees in that particular group. 
Thus, more exploration opportunities can be allocated to the root trees that have not found a path. 
Once all robots have established their respective paths, all trees involved stop growing.

\subsection{Path optimization}

\begin{figure}[htbp]
\centerline{\includegraphics{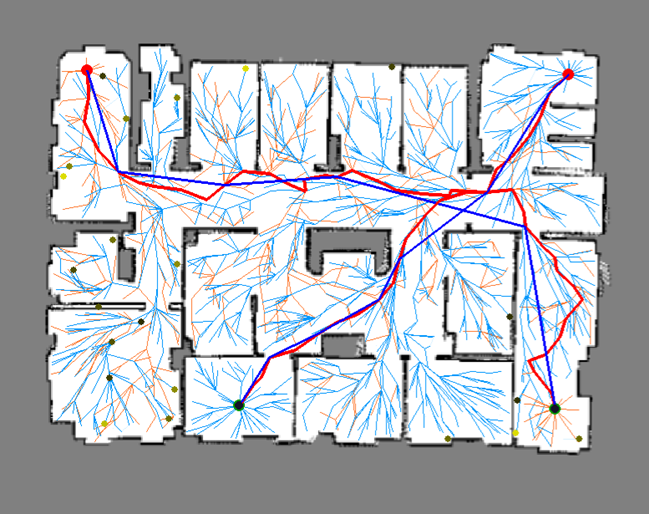}}
\caption{Example of LoS algorithm. The bold red line represents the initial path generated, while the bold blue line depicts the path optimized using the LoS algorithm. As illustrated in the figure, the optimized path exhibits substantially reduced redundant information compared to the initial path, resulting in a smoother path and decreased path length.}
\label{LoS}
\end{figure}

After finding a path, the algorithm records the position information of all nodes along the path in a list. 
However, due to the stochastic nature of node sampling, the optimality of the path cannot be guaranteed, and there may be redundant positions leading to a less smooth path. 
To enhance the quality of the generated path, this study incorporates a Line-of-Sight (LoS) checking algorithm after path generation \cite{chi2018risk}. 
The LoS algorithm can assess whether two nodes can be rewired with less cost. 
In other words, if there are no obstacles on the straight path between two arbitrary nodes, the straight path is the optimal path connecting them. 
The LoS algorithm aims to minimize costs and eliminate redundant nodes. 
This leads to a smoother and shorter path. 
The algorithmic details are presented in the following section.

Starting from the start (i.e., the initial position), the algorithm designates the initial position as the current position, denoted as $p_{tmp}$. 
It examines each subsequent position coordinate $p_{check}$ in the list to determine whether a direct line connection between $p_{tmp}$ and $p_{check}$ would result in collisions with obstacles. 
Through this examination, the algorithm identifies the first coordinate after $p_{tmp}$ that conflicts with an obstacle, and the coordinate preceding it represents the farthest non-conflicting position from $p_{tmp}$ (referred to as $p_{next}$). 
By removing all position coordinates between $p_{tmp}$ and $p_{next}$ in the list, a direct connection is established between $p_{tmp}$ and $p_{next}$, eliminating redundant positions. 
Subsequently, $p_{next}$ is set as the new current position ($p_{tmp}$), and the process continues by examining the remaining position coordinates. 
These steps are repeated until the goal position is reached (i.e., $p_{next}$ is at the goal position). 
By utilizing this LoS checking algorithm, the generated path is optimized. 
An example of the LoS algorithm is shown in Fig.~\ref{LoS}.

\subsection{Local collision avoidance}

After addressing the global path planning problem, the issue of local obstacle avoidance between robots needs to be solved. 
In this algorithm, once a path is found for a group of start and goal, the corresponding robot starts its movement toward the destination. 
However, ensuring the robots are collision-free during their motion is essential. Therefore, the algorithm incorporates the Reciprocal Velocity Obstacles (RVO) algorithm as the robot's local obstacle avoidance scheduler. 
In general, the RVO algorithm implements collision avoidance by calculating safe velocities based on other robots' current velocities and positions \cite{van2008reciprocal}. 
In RVO, a set of velocity obstacles is computed for each robot, representing a range of velocities that could result in a collision if other robots maintain their velocity. 
By considering the relative velocity obstacles, the algorithm determines the safe velocity for each robot (the safe velocity refers to the velocity outside the relative velocity obstacles). 
This effectively prevents potential collisions between robots. 
Additionally, the algorithm considers each robot's preferred velocity, which represents the desired movement velocity in the absence of other robots. 
In MA-RRdT*, the safe velocity that is closest to the robot's preferred velocity is selected as its actual velocity. 
The RVO algorithm has demonstrated effectiveness in various scenarios, including crowded environments. 
Therefore, in this study, the RVO algorithm is utilized as the local obstacle avoidance module of the robots.

\begin{algorithm}
    \caption{MA-RRdT*}
    \label{MA_RRdT_a}
    \begin{algorithmic}[1]
        \REQUIRE $n$ start points, $n$ goal points
        \STATE Initialize $2n$ root trees and $k$ disjointed trees
        \WHILE{True}
            \IF{all robots find their paths}
                \STATE Pass
            \ELSE
                \STATE Restart the disjointed trees with low probability values
                \STATE $t \gets$ Randomly pick a tree
                \STATE $q_{new} \gets$ $t$ samples
                \STATE Merge $t$ with other trees within $\epsilon$-distance of $q_{new}$
                \STATE Optimize the newly found path
                \IF{$t$ is root tree}
                    \STATE Merge $t$ with the inactive disjointed trees within $\epsilon$-distance of $q_{new}$
                \ENDIF
                \STATE Update $t$ probability
            \ENDIF
            \STATE Robots that have found their paths move one step
        \ENDWHILE
    \end{algorithmic}
\end{algorithm}

\subsection{Implementation}\label{SCM}

\begin{figure}[htbp]
\centerline{\includegraphics{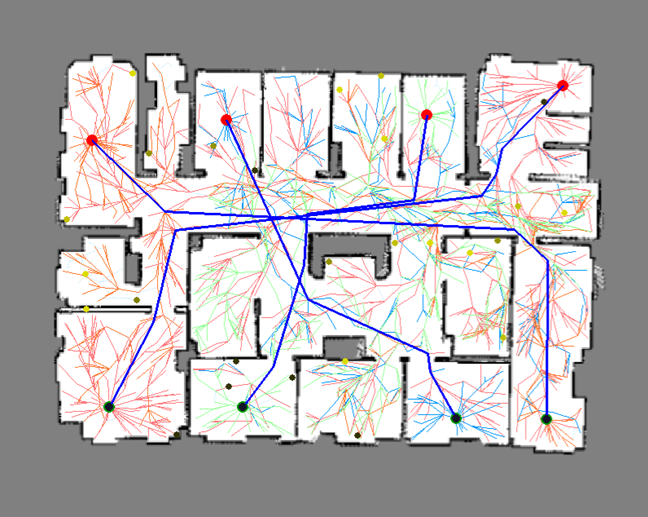}}
\caption{Example of the MA-RRdT* algorithm when all robots have reached their respective goal points. In this map, the red dots denote the start points, the green dots denote the goal points, and the black dots (superimposed on the green dots) represent the robots. The bold blue lines indicate the paths found by the algorithm, while the thin lines of varying colors correspond to the root trees of different groups. Each group exhibits a consistent color scheme for its two root trees, while distinct colors are assigned to the root trees of different groups.}
\label{MA_RRdT}
\end{figure}

The proposed algorithm improves the RRdT* single-robot path planning algorithm to address the multi-robot path planning problem. 
Based on the RRdT* algorithm, MA-RRdT* adds the number of starts and goals (corresponding to the number of robots) and generates a dedicated root tree for each start and goal. 
To ensure computational efficiency and exploration effectiveness, the position information of the disjointed trees in the space is shared among the root trees within each group. 
Moreover, a heuristic method is introduced to accelerate the exploration process for distant root trees. 
Additionally, the LoS algorithm is utilized to optimize the generated paths. 
The processed paths have a shorter length and become smoother. 
Once the paths are generated and optimized, the corresponding robots start their motion accordingly. 
The RVO algorithm is utilized as the local obstacle avoidance module to ensure safe movement. 
Algorithm~\ref{MA_RRdT_a} shows the MA-RRdT* algorithm. 
Compared to RRdT* algorithm (algorithm~\ref{RRdT_a}), it can be found that the sharing of trees and the LoS method are added to enhance the global path planning process. Moreover, the robot motion module is added.  
An example of the MA-RRdT* algorithm is shown in Fig.~\ref{MA_RRdT}. 
It can be observed that all the robots have found their corresponding paths and reached their respective goals.

\section{EXPERIMENTAL RESULTS}

\begin{figure}[htbp]
	\centering
	\subfigure[Blank] {\includegraphics[width=0.241\textwidth]{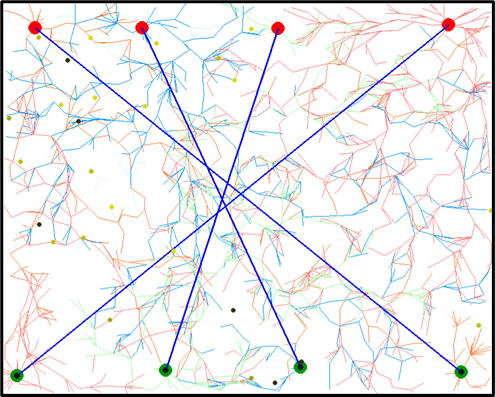}}
	\subfigure[Dense] {\includegraphics[width=0.209\textwidth]{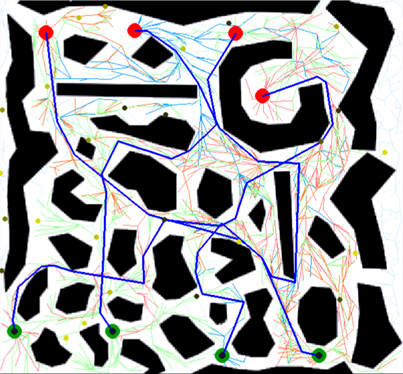}}
	\subfigure[Maze] {\includegraphics[width=0.197\textwidth]{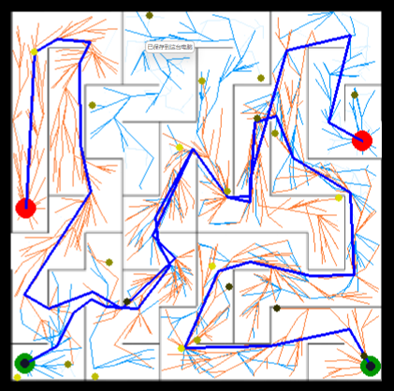}}
    \subfigure[Room] {\includegraphics[width=0.253\textwidth]{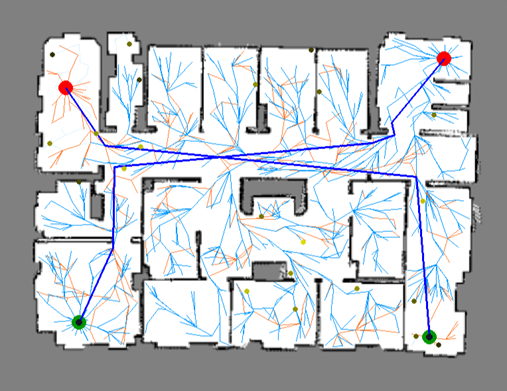}}
	\caption{Example of MA-RRdT* algorithm in different kinds of maps.}
	\label{maps}
\end{figure}

\begin{figure}[htbp]
\flushright
\subfigure{
	\rotatebox{90}{\scriptsize{~~~~~~~~~Nodes~~~~~~Path length~~~~Run time (s)~~~Plan time (s)}}
	\includegraphics[width=1\linewidth]{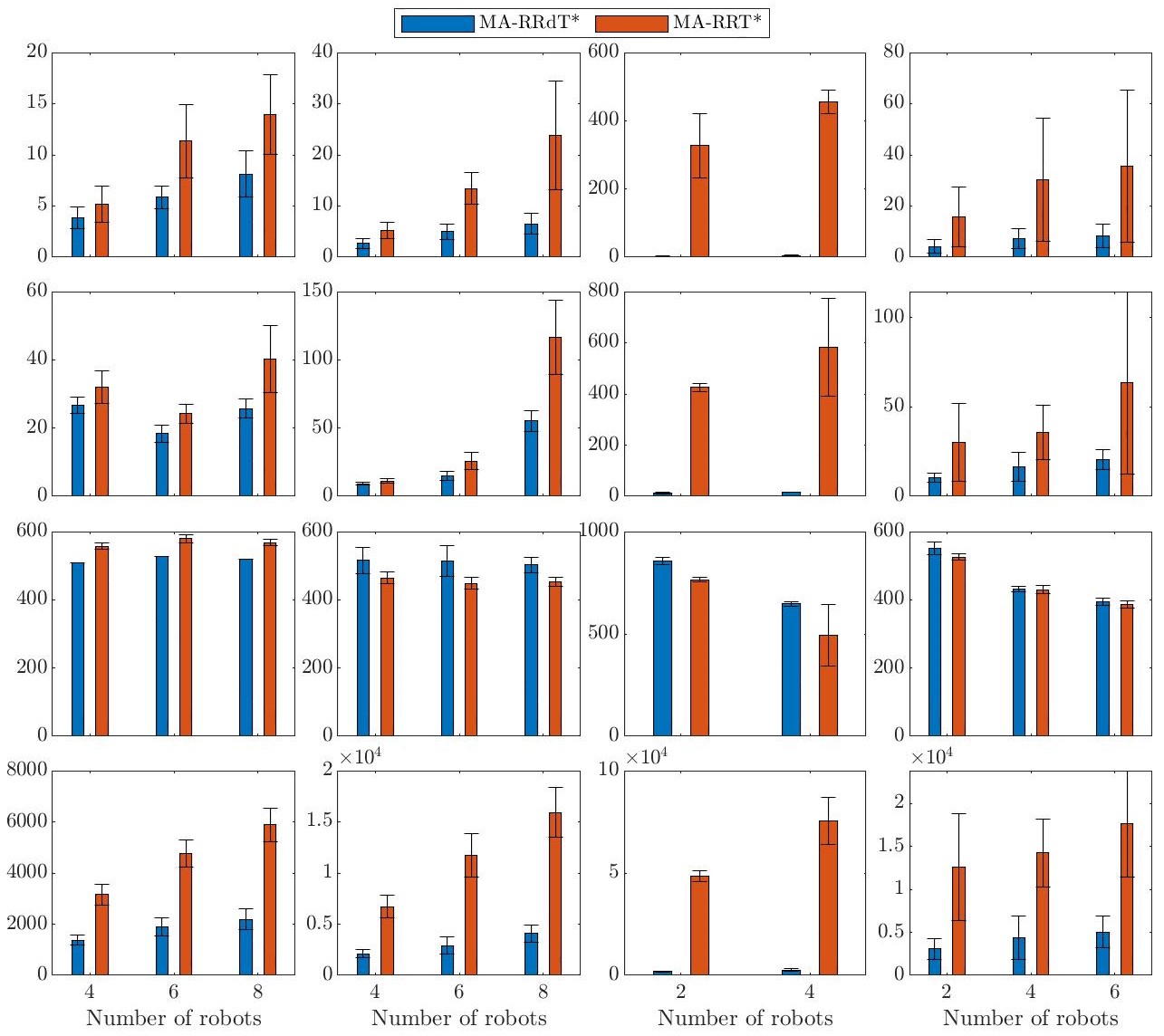}}
    \subfigure{
	\scriptsize{(a) Blank~~~~~~~~~~(b) Dense~~~~~~~~~(c) Maze~~~~~~~~~~(d) Room}}
\caption{Experimental results.}
\label{data}
\end{figure}

This work completes a comparison experiment between MA-RRdT* and the classical Multi-Agent Path Planning RRT* (MA-RRT*) algorithm. 
In MA-RRT*, each robot generates a feasible path through the RRT* method. 
The experiment involves four distinct kinds of maps: a blank map, a dense map (i.e., a map with irregular obstacles), a maze map, and a room map. 
These four maps are shown in Fig.~\ref{maps}. 
The study performs experiments with varying numbers of robots for each map to investigate the algorithm's performance and efficiency thoroughly. 
And in each scenario, this research conducted 20 runs and computed the final average plan time, run time, path length, and total number of nodes (3 runs for the maze map, as the experimental results of the two algorithms vary greatly in the maze map). 
It is noted that plan time refers to the time taken to plan the corresponding paths for each group without robot movement, while run time represents the total time with robot movement. 
The path length is obtained by dividing the total path length by the number of robots, providing the average path length for a single robot. 
The total number of nodes encompasses all sampled points, including valid and invalid ones. 
Valid nodes are the nodes that are successfully added to the tree during sampling, while invalid nodes are discarded due to collision with obstacles. 
The resulting experimental data is illustrated in Fig.~\ref{data}, with the blue part representing the MA-RRdT* algorithm and the red part representing the MA-RRT* algorithm.

In the comparison experiments, MA-RRdT* outperforms MA-RRT* in terms of plan time, run time, path length, and node count in the blank map. 
In the dense map, MA-RRdT* performs better in plan time, run time, and node count. However, it has an average path length of approximately 60 pixels longer than MA-RRT*. This slight increase in path length for MA-RRdT* can be attributed to its optimization process. 
Both algorithms progressively optimize the path during the running process, but due to shorter computation time and fewer nodes, the optimization process of MA-RRdT* is less than that of MA-RRT* (as the latter has more nodes for path optimization). 
Conversely, MA-RRdT* generates shorter paths in blank maps due to the absence of obstacles. 
And the LoS algorithm optimizes paths as straight lines from the start to the goal points in the blank map. 
In the maze map, the performance of MA-RRdT* is significantly better than MA-RRT* due to the complexity of space. 
The superior performance is attributed to two main reasons. 
Firstly, MA-RRdT* explores space more effectively by utilizing multiple trees to cover the area evenly. 
Secondly, the sampling process records previous successful information through chained directed sampling. 
These reasons enable MA-RRdT* to explore space with fewer nodes effectively and achieve quicker path planning. 
Additionally, in the room map, MA-RRdT* demonstrates shorter plan time, run time, and fewer node counts compared to MA-RRT*, while maintaining similar path lengths. Moreover, the results of MA-RRdT* are more stable.

From the above experimental results, it can be observed that the MA-RRdT* algorithm performs better than the MA-RRT* algorithm in terms of plan time, run time, and node count on each map, with the difference being more evident in complex maps. 
In the dense map, the maze, and the room map, the path length of MA-RRdT* is slightly longer than that of MA-RRT*. 
This is because MA-RRdT* uses fewer nodes and a shorter time in the tree exploration process, resulting in less path optimization than MA-RRT*. 
The distribution of nodes in both algorithms also contributes to this result. 
The nodes of MA-RRdT* are more evenly distributed in space (which also explains its ability to find paths with fewer nodes and in less time), while a large number of nodes in MA-RRT* is concentrated in certain areas, allowing MA-RRT* to optimize its path with a large number of nodes. 
As for the blank map, the shorter path length of MA-RRdT* is due to the effectiveness of the LoS algorithm in the absence of obstacles, making MA-RRdT* generates shorter paths than MA-RRT*. 
In conclusion, the MA-RRdT* algorithm enables robots to complete global path planning using fewer sampling nodes in a shorter time, especially in maps with narrow passages such as mazes and rooms. 
Although its path length is slightly longer than that of the MA-RRT* algorithm in complex maps, this sacrifice is acceptable to achieve higher planning efficiency.

\section{CONCLUSIONS AND FUTURE WORK}
In conclusion, this article introduces a novel multi-robot path planning algorithm, MA-RRdT*, based on the single-robot path planning algorithm RRdT*. 
To address multi-robot path planning tasks, this algorithm improves the RRdT* by providing spatial information obtained from randomly distributed disjointed trees for each group of start and goal. 
It aims to enable the planning of feasible paths for each robot to its corresponding goal point by using a small number of sampling nodes during a short time. 
After implementing the MA-RRdT* algorithm, this work completes comparative experiments to demonstrate and validate its effectiveness in terms of time and node count.

Possible avenues for future work include using neural networks to perceive the environment to provide prior knowledge for the algorithm implementation \cite{wang2020neural}, and using real-world applications to further evaluate and improve the proposed algorithm \cite{xia2023collaborative}.

%
\bibliographystyle{IEEEtran}
\bibliography{ref} 

\end{document}